\documentclass[iicol,sn-basic]{sn-jnl} 
\usepackage{graphicx}%
\usepackage{multirow}%
\usepackage{amsmath,amssymb,amsfonts}%
\usepackage{amsthm}%
\usepackage{mathrsfs}%
\usepackage[title]{appendix}%
\usepackage{xcolor}%
\usepackage{textcomp}%
\usepackage{manyfoot}%
\usepackage{booktabs}%
\usepackage{algorithm}%
\usepackage{algorithmicx}%
\usepackage{algpseudocode}%
\usepackage{listings}%
\newcommand{\mathbbm}[1]{\text{\usefont{U}{bbm}{m}{n}#1}}
\theoremstyle{thmstyleone}%
\theoremstyle{thmstyletwo}%

\theoremstyle{thmstylethree}%

\begin{document}

\title[Article Title]{Neuromorphic Valence and Arousal Estimation}


\author*[1]{\fnm{Lorenzo} \sur{Berlincioni}}\email{lorenzo.berlincioni@unifi.it}
\equalcont{These authors contributed equally to this work.}

\author*[1]{\fnm{Luca} \sur{Cultrera}}\email{luca.cultrera@unifi.it}
\equalcont{These authors contributed equally to this work.}

\author*[2]{\fnm{Federico} \sur{Becattini}}\email{federico.becattini@unisi.it}

\author[1]{\fnm{Alberto} \sur{Del Bimbo}}\email{alberto.delbimbo@unifi.it}

\affil*[1]{\orgdiv{MICC}, \orgname{University of Florence}, \orgaddress{\street{Viale Morgagni 65}, \city{Florence}, \postcode{50134}, \country{Italy}}}

\affil[2]{\orgname{University of Siena}, \orgaddress{\street{Via Roma 56}, \city{Siena}, \postcode{53100}, \country{Italy}}}


\abstract{
Recognizing faces and their underlying emotions is an important aspect of biometrics. In fact, estimating emotional states from faces has been tackled from several angles in the literature. In this paper, we follow the novel route of using neuromorphic data to predict valence and arousal values from faces. Due to the difficulty of gathering event-based annotated videos, we leverage an event camera simulator to create the neuromorphic counterpart of an existing RGB dataset. We demonstrate that not only training models on simulated data can still yield state-of-the-art results in valence-arousal estimation, but also that our trained models can be directly applied to real data without further training to address the downstream task of emotion recognition. In the paper we propose several alternative models to solve the task, both frame-based and video-based.}

\keywords{valence, arousal, event camera, face analysis, neuromorphic vision}



\maketitle

\begin{figure}[t]%
\centering
\includegraphics[width=0.85\columnwidth]{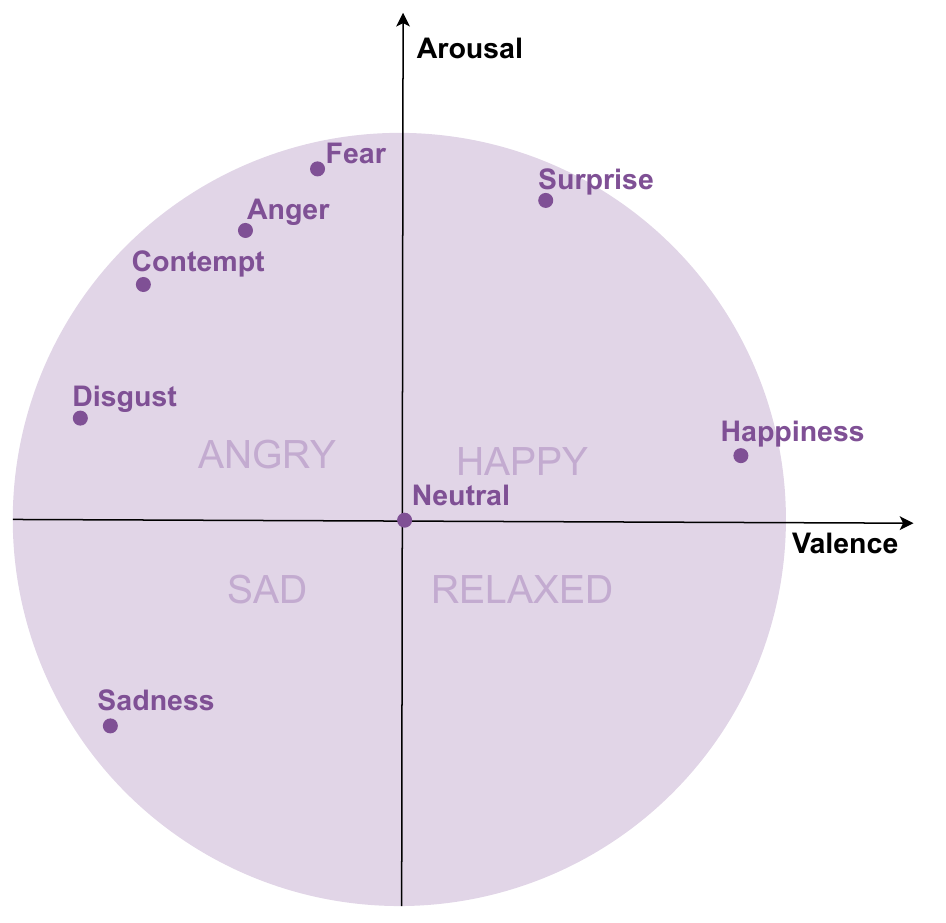}
\caption{\textbf{Valence}-\textbf{Arousal} unit circle. Values can be directly mapped into emotions \cite{mikels2005emotional}.}\label{figar}
\end{figure}

\section{Introduction}\label{sec1}

Analyzing humans and their behaviors is one of the most important fields of artificial intelligence and computer vision. Such importance stems from the repercussions that a technology capable of understanding humans can have on society: being able to recognize an individual is fundamental for security; analyzing biometrics offers intriguing possibilities for patient monitoring in healthcare;  understanding behaviors and emotions enables smart human-robot collaborations in private spaces as well as in industry. In synthesis, human understanding is revolutionizing our society and our behaviors, both in our private sphere and in workspaces, where humans and AI-driven robotic agents are starting to work alongside.
To ensure a seamless interaction in this sense though, recognizing individuals and their behaviors is not enough. Robotic agents, let them be actual humanoid robots, vision-based software modules or conversational agents, must infer the mood of the human they are observing so to provide a more natural way of interacting as well as to better plan an appropriate reaction that ensures safety and an harmonious work environment.
%\subsection{Valence and Arousal}\label{subsec2}

\begin{figure*}[t]%
\centering
\resizebox{0.85\textwidth}{!}{
\includegraphics[width=\textwidth]{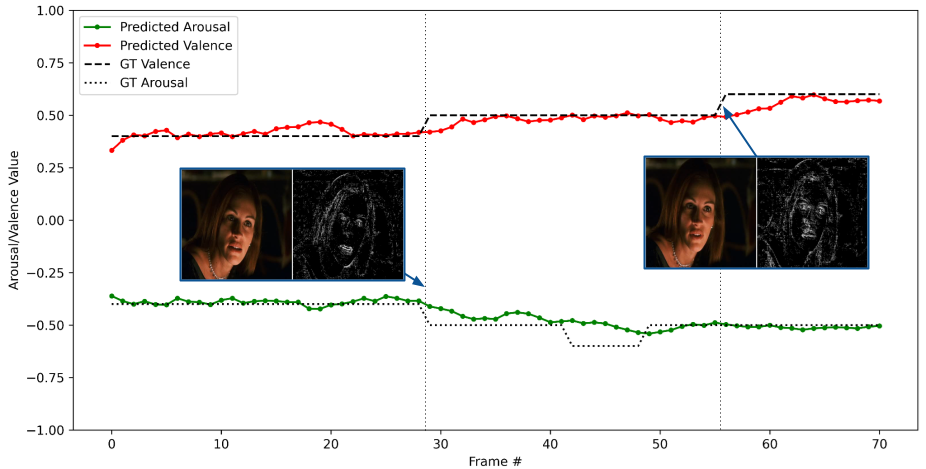}}
\caption{Illustration of RGB and event frames in a sample video over its relative valence and arousal plot.}\label{fig:timeframes}
\end{figure*}

Therefore, in this paper, we focus on analyzing faces in order to estimate human moods and emotions. A lot of prior work exists in this field, mostly focusing on analyzing facial expressions to understand the underlying emotion.
Several works in the field of expression recognition focused on detecting facial action units \cite{ekman1978facial,Rudovic2015,Kaltwang2015} or they formulated the problem as close-set classification task over a limited number of emotions. A different, more recent, approach instead poses the problem as a regression task over two continuous dimensions measuring positive and negative affectivity (valence) and the level of excitement of the expressed emotion (arousal) \cite{b764d6ea581b44beb132a7b8669b7eea,GUNES2013120}.

%Emotional experiences can be described by two terms:
%VALENCE: a scale that measures the positive or negative affectivity (so the pleasantness or unpleasantness of something)
%AROUSAL: is a measure of how calming or exciting the information is.

We follow the latter approach of estimating valence and arousal, as it can provide a punctual frame-by-frame estimate of the mood in a continuous way and can then be translated into more specific interpretations such as emotion categories (Fig. \ref{figar}). However, we argue that relying on traditional RGB cameras can have limitations in processing human faces effectively. Human emotions are often manifested through fast, inconceivable and involuntary facial muscle movements, that can be completed within a few milliseconds \cite{yan2013fast}. Such movements might not even be fully observable with traditional RGB cameras. Nonetheless, for many practical applications, it is necessary to achieve a more fine-grained resolution of the continuously produced micro-movements of the human face. To address this issue, a few methods have been proposed recently that analyze faces with the use of a neuromorphic camera (often referred to as an event camera) rather than an RGB one \cite{berlincioni2023neuromorphic, becattini2022understanding, lenz2020event, ryan2023real, shariff2023neuromorphic, bissarinova2023faces}.
Unlike traditional cameras, neuromorphic sensors work asynchronously and capture events, i.e. per-pixel illumination changes, and have highly desirable properties such as microsecond latency, high-dynamic ranges and low power consumption.
Such properties enable event cameras to capture subtle variations and micro-expressions in human faces (and, therefore, emotions) at a remarkably high temporal resolution.
In addition, analyzing faces with event cameras is also favorable for preserving the privacy of the subjects. Streams are in fact less interpretable for the human eye and can be scrambled in order to make the subjects unrecognizable \cite{ahmad2023person} without altering the capacity of computer vision models.

In this paper, we present the first approach to model valence and arousal in human faces using neuromorphic data (Fig. \ref{fig:timeframes}). To address this task, we rely on an event simulator~\cite{Hu2021-v2e-cvpr-workshop-eventvision2021} capable of converting RGB videos into simulated event streams. This solution is sub-optimal compared to using real event-based videos but enables several key factors that would be hard and costly to obtain: on the one hand, it provides us with a fully labeled neuromorphic dataset, since valence-arousal annotations can be directly transferred from the original dataset; on the other hand, it allows us to train computer vision models without the need of collecting additional data with an event camera, which also entails that we do not require any manual annotation. In addition, we perform zero-shot transfer experiments onto real event data, demonstrating also that our models can be adopted for the downstream task of emotion classification without any further training.
 
%Don’t output frames but events that are single pixel intensity changes at microsecond resolution
%An event  is triggered when the magnitude of the log brightness at pixel  and time  has changed by more than a threshold since the last event at the same pixel

%Event cameras can capture illumination changes, and thus motion, at an extremely fast pace. Being able to effectively analyze faces at such a data rate would allow us to precisely characterize expressions and their underlying emotions.
%However, neuromorphic face analysis in the literature is an almost unexplored field of research. Yet a few preliminary results indicating its effectiveness over RGB exist \cite{becattini2022understanding, berlincioni2023neuromorphic}.
%Interpreting human faces is fundamental to many applications, ranging from simple detection up to more complex tasks such as emotion recognition or 3D modeling.
% Analyzing faces with RGB is supported by lots of annotated datasets and models
%Such importance has resulted in a huge amount of research in this field, which nowadays is supported by a plethora of annotated datasets and open-source models. These provide off-the-shelf tools acting as building blocks for face analysis applications such as face detection \cite{dlib09}, landmark detection \cite{bulat2017far} and gaze estimation \cite{zhang2017mpiigaze}, just to name a few.
% and suggest its applicability in several applications such as robot-human interaction \cite{becattini2022understanding}, sentiment analysis \cite{berlincioni2023neuromorphic}, driver monitoring \cite{}.
In summary, the main contributions of our paper are the following:
\begin{itemize}
	\item We investigate the problem of estimating valence and arousal from event streams. To the best of our knowledge, we are the first to address such a problem.
	
	\item We propose several deep learning solutions, proposing different frame-based and video-based architectures. To train such architectures we rely on simulated event data, obtained by converting AFEW-VA \cite{Kossaifi2017AFEWVADF}, an RGB dataset manually labeled for valence and arousal.
	
	\item We demonstrate that our models can be successfully applied also on real event streams from the NEFER dataset \cite{berlincioni2023neuromorphic} and that we can address the task of emotion estimation directly from the predicted valence and arousal values, without additional training on the new data.
\end{itemize}

\section{Previous Work}\label{sec_previous_work}

\textbf{Neuromorphic Vision}
Neuromorphic vision involves data acquisition methods based on event cameras, bio-inspired vision sensors that have been recently introduced \cite{delbruckl2016neuro, posch2014retino}. Unlike traditional vision systems, neuromorphic sensors generate asynchronous streams of events rather than a frame sequence with a predetermined frame-rate. Instead of obtaining frames from the camera, we now obtain \textit{events}, which are local changes in the brightness of a single pixel. What makes these sensors extremely interesting is the fact that events can be fired at sub-millisecond rates \cite{lichtsteiner2008asynch}.
To this day, event cameras have been applied in several domains. Of particular interest, is the possibility to enhance robots that require quick response times with onboard low-latency devices. This has aided applications such as autonomous drone navigation \cite{falanga2020dynamic}, SLAM \cite{mueggler2017event, 9813406}, tracking \cite{seok2020robbust,renner2020event} and object detection in automotive \cite{perot2020learning}.

When working with event data, a few considerations have to be taken into account. Notably, these neuromorphic sensors exhibit the distinctive characteristic of not outputting any data unless a localized change in brightness is detected, effectively conserving resources and minimizing bandwidth consumption \cite{finateu2020back, gallego2020event}. In general, the fact that events are not generated synchronously entails the need for an intermediate representation of events that can be processed, for instance, by deep learning architectures. Whereas dedicated architectures exist, such as Spiking Neural Networks \cite{barchid2023spiking}, a common way to proceed is to accumulate the events that happen in synchronous time intervals to generate frames that can be fed to a convolutional neural network. Several event aggregation strategies exist \cite{mueggler2017fast, innocenti2021temporal, nguyen2019real, cannici2020differentiable}, which are often capable of injecting some temporal context into the information contained in each pixel.

In this paper, we leverage event data for a newborn field of research, that is neuromorphic face analysis. Analyzing faces with an event camera in fact permits to capture high-frequency information that might be difficult to capture with standard cameras. For instance, facial action units are tied to extremely fast muscle movements that appear as small movements in a video \cite{yan2013fast}. Just a few works exist involving event cameras and faces.
Face detection \cite{bissarinova2023faces} and face pose estimation \cite{savran2020face} have been addressed, but also lip reading \cite{bulzomi2023end} and eye-blink detection \cite{lenz2020event}.
Among the first attempts to estimate affective information from event videos, \cite{becattini2022understanding} estimated positive or negative facial reactions when observing fashion items and \cite{berlincioni2023neuromorphic} classified 7 basic emotions. Differently from these works, we focus on estimating valence and arousal, which we belive to be a finer modeling of facial expressivity. In fact, we also experimentally demonstrate that emotions can be directly inferred from our predicted valence and arousal value, even when tested on a different dataset from the one used for training.

\textbf{Emotion estimation}\label{sec3}
Most of the research in literature on emotion estimation focused on facial expression recognition,
facial action unit detection, and expression classification \cite{savchenko2022classifying, kollias2019expression, li2022facial, schoneveld2021leveraging}.
Mikels’~Wheel of Emotions \cite{mikels2005emotional} is a visual representation of emotion classes in the valence-arousal space, a widely-used emotion model from psychology. As shown
in Fig. \ref{figar}, emotions on Mikel’s wheel are separated into
eight categories
%(i.e., amusement, anger, awe, contentment, disgust, excitement, fear, and sad)
as well as two polarities (i.e., positive and negative).
\cite{toisoul2021estimation} propose a method for real-time applications to estimate both categorical and continuous emotions.
\cite{kossaifi2020factorized} introduces CP-Higher-Order Convolution, a tensor factorization framework unifying low-rank tensor decompositions and efficient convolutional block design. Enabling higher-order transduction, the approach facilitates training on a specific domain (e.g., 2D images) and generalizing seamlessly to higher-order data like videos, demonstrating superior performance in spatio-temporal facial emotion analysis on large-scale datasets.
Different from the aforementioned works,  \cite{parameshwara2023efficient} employs a Siamese network trained with image pairs and a contrastive loss. This enables the network to estimate emotional dissimilarity and quantify valence and arousal differentials for given image pairs.
\cite{handrich2020simultaneous} use a YOLO-based model to predict face bounding boxes, basic emotions and valence-arousal values. 
\cite{mitenkova2019valence}, instead, propose a tensor-based method to predict continuous values of valence and arousal. Also \cite{kollias2020deep} introduce a data augumentation technique to train Deep Neural Network to perform valence-arousal estimation.
%The methods described so far,  propose results for valence-arousal estimation on the AFEW-VA dataset \cite{Kossaifi2017AFEWVADF}. 

On the other hand, other methods prefer a more categorical approach, aiming to predict the 8 emotions on Mikels's wheel described earlier rather than continuous valence-arousal values. \cite{wen2023distract}, for instance, proposes an approach based on multi-head attention for emotion classification, achieving remarkable results. \cite{savchenko2021facial} introduces a streamlined training approach for a lightweight CNN in facial analytics, achieving state-of-the-art results in video-based emotion analysis.
\cite{mao2023poster} combine facial landmark and image features through two-stream pyramid cross-fusion design obtaining state-of-the-art results in emotion recognition.
%Other notable works in emotion classification include: \cite{savchenko2022classifying, kollias2019expression, li2022facial, panagiotis2106exploiting, schoneveld2021leveraging}. 
Unlike the approaches described so far, in this paper, we propose to focus on event videos, a domain that has been relatively unexplored in the literature but appears to be promising, particularly in areas such as face analysis and emotion recognition.

% \section{Motivation}
% As previously stated the event domain is comparatively new with respect to other acquisition sources, this implies a real scarce amount of data which in turn might weaken the ability to compare this approach and the potential applications. We therefore aim at bridging the domain gap in order to prove the 
% \todo{
% Main contribution è VA su event che non esiste. Il problema in effetti è valorizzarlo e giustificarlo. Il solito pippone di modellare micromovimenti con l'event si può sempre fare ma il fatto di avere il simulatore di mezzo complica un po' le cose.
% Visto il topic della special issue andrà sicuramente motivato parlando di biometrics per cyber security. Potremmo dire che tramite l'uso di event camera è più facile garantire la privacy (c'è un paper di IIT che parla di privacy con event, in realtà dice che non è privacy preserving ma propone un modo per farlo diventare tale) e quindi un sistema per analizzare volti in ambito industriale con event potrebbe essere più efficace, più sicuro e più veloce (generiamo meno frame dell'rgb) e magari funzionare anche in condizioni avverse (solite qualità dell'event)}

% Event cameras can capture illumination changes, and thus motion, at an extremely fast pace. Being able to effectively analyze faces at such a data rate would allow us to precisely characterize expressions and their underlying emotions.
% However, neuromorphic face analysis in the literature is an almost unexplored field of research.

\section{Simulating Neuromorphic Data}\label{sec:simulating}

Training a computer vision model based on neuromorphic streams is not straightforward. The main challenge that has to be faced is the lack of data sources from where to obtain meaningful samples. Videos cannot be crawled from the web and new datasets need to be recorded and labeled from scratch. Automatizing such pipeline is not trivial as off-the-shelf traditional computer vision models (e.g. face detectors) are ineffective on event frames. The intrinsic structure of the data itself makes it hard to annotate it since when no illumination change is detected by the sensor no signal is produced.

Luckily, event camera simulators have been proposed in the literature, namely, ESIM \cite{Rebecq18corl} and V2E \cite{Hu2021-v2e-cvpr-workshop-eventvision2021}. These simulators are capable of producing neuromorphic counterparts from RGB videos. To this end, they first perform a temporal upsampling of RGB frames, with a rate that adapts to the video content and its estimated visual dynamics (the more the video changes, the more frames are added). Then, synthetic events are generated by analyzing the differences between adjacent frames.

In this paper, we adopt V2E \cite{Hu2021-v2e-cvpr-workshop-eventvision2021} to convert an RGB dataset labeled with valence and arousal values for each frame. In particular, we use the AFEW-VA dataset \cite{Kossaifi2017AFEWVADF}, which consists of a collection of 600 RGB videos extracted from movies.
Each per-frame annotation is a discrete value in the range of -10 to 10. Along with these annotations, the positions of 68 facial landmarks are also provided.
Videos range from around 10 frames to longer clips (more than 120 frames); in total, there are 30,000 frames in the entire dataset.
%In order to obtain event stream we use as simulated pipeline as explained in Section \ref{subsection_simulated}.

Once the videos are converted, we need to map the annotations onto event data. To do so, we assign to each annotation a timestamp corresponding to the one of the frame within the video. When we generate event frames (see Sec. \ref{sec4}) we then label them with valence and arousal by looking for the annotation with the closest timestamp to the average timestamp of the events in the neuromorphic frame.

In the following, we will outline our training pipeline for learning to predict valence and arousal from the simulated event streams, both leveraging frame-based models as well as video-based models.
Interestingly, our experimental validation shows that we are able to obtain state-of-the-art results on the AFEW-VA dataset when comparing our results with RGB-based models from the literature. We also show that our trained models can be directly applied on real event data, demonstrating excellent zero-shot transfer capabilities on the related task of emotion recognition on the NEFER dataset \cite{berlincioni2023neuromorphic}.

\section{Valence and Arousal Estimation}\label{sec4}

Given a video sequence $ v = \{f_0, f_1, ..., f_{T-1}\}$ of $T$ frames, our goal is to regress a pair of valence and arousal values ($\hat{v}_i, \hat{a}_i$) for each frame, so to match the correspondent ground truth values ($v^*_i, a^*_i$) with $i=0,...,T-1$.
The problem can be addressed by analyzing single frames or sequences of frames, thus providing a temporal context to the prediction. In the following, we will present several alternative models for predicting valence and arousal from both frames and video chunks.
In both cases, the methods we propose all leverage frame-based representations of events. Neuromorphic data, in fact, is natively represented as a list of asynchronous events, yet it is common practice to aggregate events into frames by gathering all the activations that happen within an aggregation time $\Delta t$ \cite{mueggler2017fast, innocenti2021temporal, nguyen2019real}. 
This allows us to use standard computer vision models such as convolutional neural networks even with neuromorphic data.

In particular, we choose to represent events with the Temporal Binary Representation (TBR) \cite{innocenti2021temporal} strategy.
%TBR is an event aggregation strategy to convert the output stream of an event camera into more classical output image frames.
To compute the TBR frame encoding we do the following. After setting a fixed accumulation time $\Delta t$, we can build a binary representation of the frame $b$ by checking for the presence of any event at each location $(x,y)$, that is $b_{x,y} = \mathbbm{1}(x,y)$, where $\mathbbm{1}$ is an indicator function that is equal to 1 is an event is present in position $(x,y)$ during the accumulation interval and 0 otherwise.

Once the binary representation has been created, it is possible to collect N consecutive frames and concatenate them together as a tensor $B \in \mathbb{R}^{H \times W \times N}$, where $W$ and $H$ are respectively the width and the height of the frame. This yields for each pixel a binary string $B_{x,y} = [b^{0}_{x,y},b^{1}_{x,y}, ... , b^{N-1}_{x,y}]$ that can be converted to a scalar through a binary-to-decimal conversion. By doing so, TBR manages to create a frame processable by traditional computer vision pipelines along with the benefit of retaining temporal information spanning across a time interval of $N \times \Delta t$ within the value of each pixel while needing a minimal memory footprint.
In our experiments, we used $\Delta t = 5 $ milliseconds and $N = \{8, 16\}$.
%\todo{The proposed method first generates sequences of intermediate binary representations, which are then losslessly transformed into a compact format by simply applying a binary-to-decimal conversion. This strategy allows us to encode temporal information directly into pixel values.}

\subsection{Models}
\label{sec:model}
We follow two main protocols and therefore two main families of models.
The \textit{frame-based} ones are those models working with a single frame, which is they have a single $(v,a)$ output given a single event-frame input.
The \textit{video-based} models instead work at video level, having an output per \textit{event frame} and a sequence of frames as an input.
For the frame-based models we utilize two architectures to address the task: ResNet18 \cite{he2016deep} and Vision Transformer (ViT) \cite{dosovitskiy2020image}. For the Vision Transformer configuration, we employ four attention heads with a depth of 4, utilizing patch sizes of 8. Across both ViT and ResNet, we maintain consistent input dimensions of $224 \times 224$ pixels.
For the video-based models, we adopt four distinct architectures: IC3D, ResNet+LSTM, ResNet+Transformer and a custom architecture that we refer to as ResNet+Fusion.
Both the ResNet+LSTM and ResNet+Transformer models utilize a pretrained ResNet18 on ImageNet, extracting features of dimension 1024. During the training phase, ResNet is kept frozen for both models. In the first case, the output features from ResNet are fed into a sequence of 3 LSTM layers with a hidden size of 256. In the second case, the features are processed by a transformer-based architecture with 4 heads, 6 encoders, and 6 decoders. Both models employ a final MLP (comprising two layers) for regressing valence and arousal values for each frame of the input sequence. 
Conversely, the ResNet+Fusion model employs an unfrozen ResNet18 during training. The resulting 1024-dimensional output features from ResNet are then directed into two distinct heads. The first head processes video-level features by stacking all the frame features together, while the second head handles frame-level features individually. Both heads generate 128-dimensional features using multiple linear layers. Subsequently, the features extracted from both heads are concatenated, and a final MLP, consisting of two linear layers, predicts valence and arousal values for each frame in the sequence. Significantly, this model excels in learning features at both video and frame levels, thereby enhancing its ability to discern subtle patterns throughout the entire sequence. Since we process several frames at a time, we have to fix the sequence length. In our experiments, we process chunks of 6 frames individually.
Lastly, IC3D employs an architecture inspired by Inception3D \cite{carreira2017quo}. However, unlike the approach in \cite{carreira2017quo}, we utilize a single data stream, resulting in a single-branch architecture composed of 3D convolutions.
The activation function employed in all MLPs across the models is ReLU. The models were trained using the AdamW optimizer with an initial learning rate of 0.0001. For every listed model we also employ a scheduler that halves the learning rate every 50 epochs.

\section{Experiments}\label{sec_exps}
In this section, we define our experimental methodology and showcase the primary outcomes of the proposed approach.
We present the results of our simulated-data pipeline in terms of a valence-arousal regression task over the AFEW-VA dataset \cite{Kossaifi2017AFEWVADF, toisoul2021estimation}, also by comparing the results with RGB baselines from the literature. We then demonstrate the zero-shot transfer capabilities of our models on a related downstream task using real event videos, i.e. emotion classification on the recently proposed NEFER dataset \cite{berlincioni2023neuromorphic}.

We only train our models on the synthetically generated event videos obtained by applying the V2E simulator of the AFEW-VA (as presented in Sec. \ref{sec:simulating}). The evaluation is then performed on AFEW-VA by following the experimental validation protocol of prior works known as subject-independent \cite{Kossaifi2017AFEWVADF} and on NEFER by using the test split provided by the authors.

Here we first introduce the metrics used to evaluate our models, then we present the results and perform an ablation study on the TBR encoding strategy, varying the number of bits $N$ used in the data representation scheme.

\subsection{Metrics}
We employ multiple metrics for performance evaluation over both AFEW-VA and NEFER datasets.
Given that $y^*$ and $\hat{y}$ represent the ground truth and the predicted values, we can define several metrics to evaluate the different models.
On AFEW-VA we adopt the following ones:
\begin{itemize}
\item Root Mean Square Error (RMSE) evaluates how close predicted values are from the target values:
\begin{equation}
    RMSE(y^*,\hat{y}) = \sqrt{\mathbb{E}(y^*-\hat{y})^2}
\end{equation}
\item Pearson Correlation Coefficient (PCC) measures how correlated predictions and target values are:
\begin{equation}
    PCC(y^*,\hat{y}) = \frac{\mathbb{E}(y^*-\mu_{y^*})(\hat{y}-\mu_{\hat{y}})}{\sigma_{y^*}\sigma_{\hat{y}}}
\end{equation}
\item Sign Agreement (SAGR) is a measure to evaluate if the sign of the predicted value matches with the target.
\begin{equation}
    SAGR(y^*,\hat{y}) = \frac{1}{n}\sum_{i=1}^{n}\delta(sign(y^*_{i}),sign(\hat{y}_{i}))
\end{equation}
\end{itemize}

\begin{table}[t]
	\centering
	\resizebox{\linewidth}{!}{
		\begin{tabular}{lcccccc}
			\toprule
			Model & \multicolumn{3}{c}{Arousal} & \multicolumn{3}{c}{Valence}\\
                & RMSE$\downarrow$ & PCC$\uparrow$ & SAGR$\uparrow$
                & RMSE$\downarrow$ & PCC$\uparrow$ & SAGR$\uparrow$ \\
			\midrule
			ResNet18 & 0.200 & 0.307 & \textbf{0.803} & 0.246 & \textbf{0.110} & 0.\textbf{466} \\
			ViT & \textbf{0.173} & \textbf{0.340} & 0.802 & \textbf{0.211} & 0.005 & 0.463	\\
			%?? & ?? & ?? & ?? & ?? & ?? & ??\\
			\bottomrule
		\end{tabular}
	}
	\caption{Result on the AFEW-VA event dataset for frame-based models.}
	\label{tab:AFEW_frame}
\end{table}

\begin{table}[t]
	\centering
	\resizebox{\linewidth}{!}{
		\begin{tabular}{lcccccc}
			\toprule
			Model & \multicolumn{3}{c}{Arousal} & \multicolumn{3}{c}{Valence}\\
                & RMSE$\downarrow$ & PCC$\uparrow$ & SAGR$\uparrow$
                & RMSE$\downarrow$ & PCC$\uparrow$ & SAGR$\uparrow$ \\
			\midrule
			ResNet+Fusion  & \textbf{0.124} & \textbf{0.580} & 0.805          & \textbf{0.191} & \textbf{0.297} &	0.451 \\
			IC3D        & 0.130          & 0.520          & 0.812 & 0.201          & 0.141 & \textbf{0.455}	\\
			ResNet+LSTM & 0.153          & 0.525          & 0.799          & 0.222          & 0.209 & 0.413 \\
            ResNet+Transf. & 0.133          & 0.490          & \textbf{0.815}          & 0.226          & 0.132 & 0.449 \\
			\bottomrule
		\end{tabular}
	}
	\caption{Result on the AFEW-VA event dataset for video-based models.}
	\label{tab:AFEW_video}
\end{table}

\subsection{Results}
In Tab. \ref{tab:AFEW_frame} and Tab. \ref{tab:AFEW_video}, a comparison between the proposed models for both the \textit{frame-based} and \textit{video-based} approaches is provided.
%In both tables, the metrics used are: RMSE, PCC, and SAGR.
Regarding \textit{frame-based} models, ViT achieves the most interesting results, yet both models achieve an RMSE lower or equal to 0.2. To correctly interpret these results, it has to be noted that we represent valence and arousal values in the range [-1, 1], as commonly done for evaluation \cite{toisoul2021estimation, Kossaifi2017AFEWVADF}.

As for \textit{video-based} models, ResNet+Fusion stands out. Notably, ResNet+Fusion also emerges as the model with the overall best performance among all the approaches. Moreover, all video-based methods perform better than models trained to analyze just a single frame. This suggests that in order to predict valence and arousal effectively, providing a temporal context can be helpful. We believe that providing a temporal context also reduces the chance of having frames with low content due to lack of movement in the video (and therefore lack of events).
%The superior performance of ResNet18+cat underscores the effectiveness of capturing temporal dependencies across frames for enhanced predictive capabilities.

Interestingly, for all methods, arousal appears to be easier than valence. This trend is also confirmed by prior works, as shown in Tab. \ref{tab:afew_comparison}. Here, we compare our best frame-based model (ResNet+Fusion) and our best video-based model (ViT) with state-of-the-art methods trained and tested on the original RGB version of the dataset.
Both methods are capable of performing better or on par compared to prior works. This points towards the direction, also suggested by other works in the neuromorphic literature \cite{becattini2022understanding, berlincioni2023neuromorphic}, that event-based representations might help models to focus more on informative content, filtering out distractors such as background and textures that can interfere with the learning process.
%As highlighted in table \ref{tab:afew_comparison}, our approach  (ResNet18+cat) significantly outperforms the state-of-the-art for both RMSE valence and arousal metrics.

\begin{figure*}[t]%
\centering
\includegraphics[width=\columnwidth]{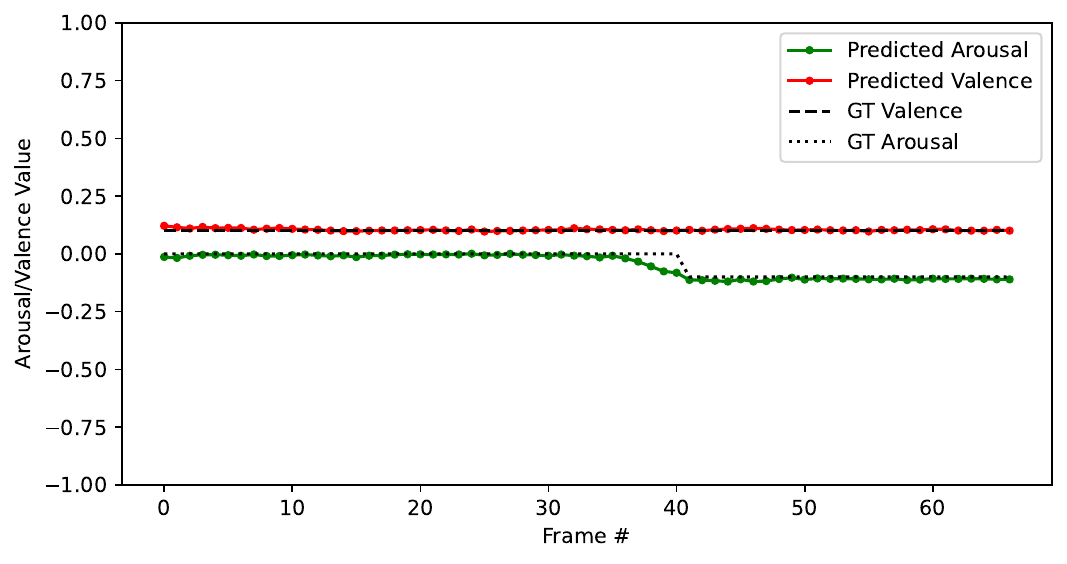}
\includegraphics[width=\columnwidth]{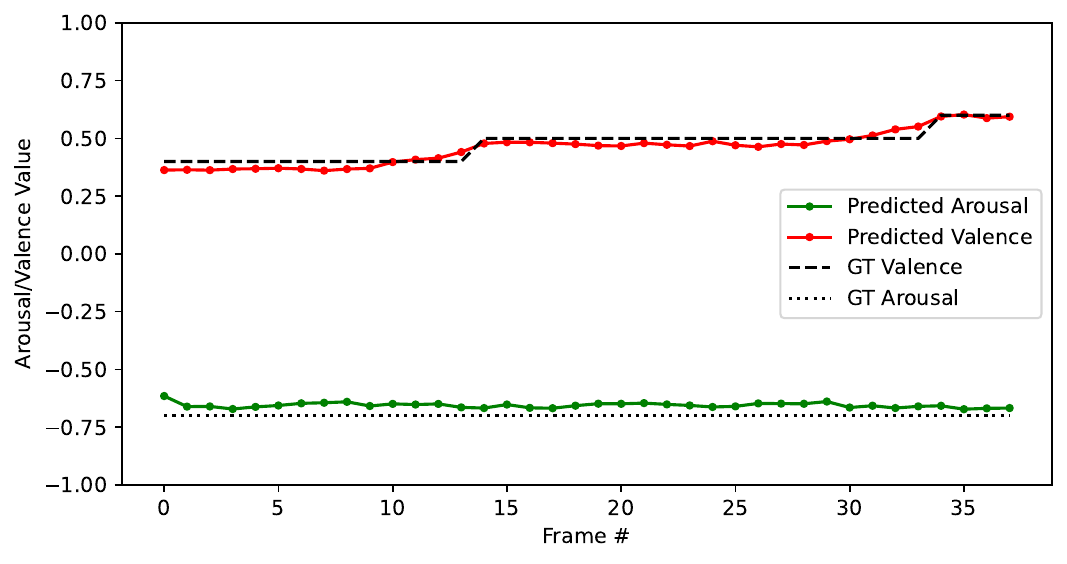}
\includegraphics[width=\columnwidth]{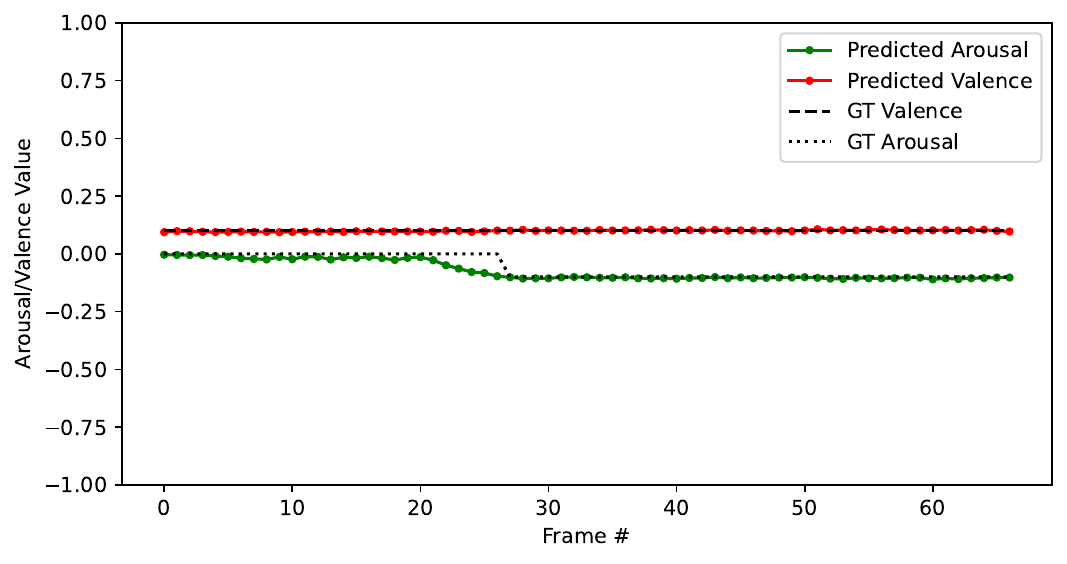}
\includegraphics[width=\columnwidth]{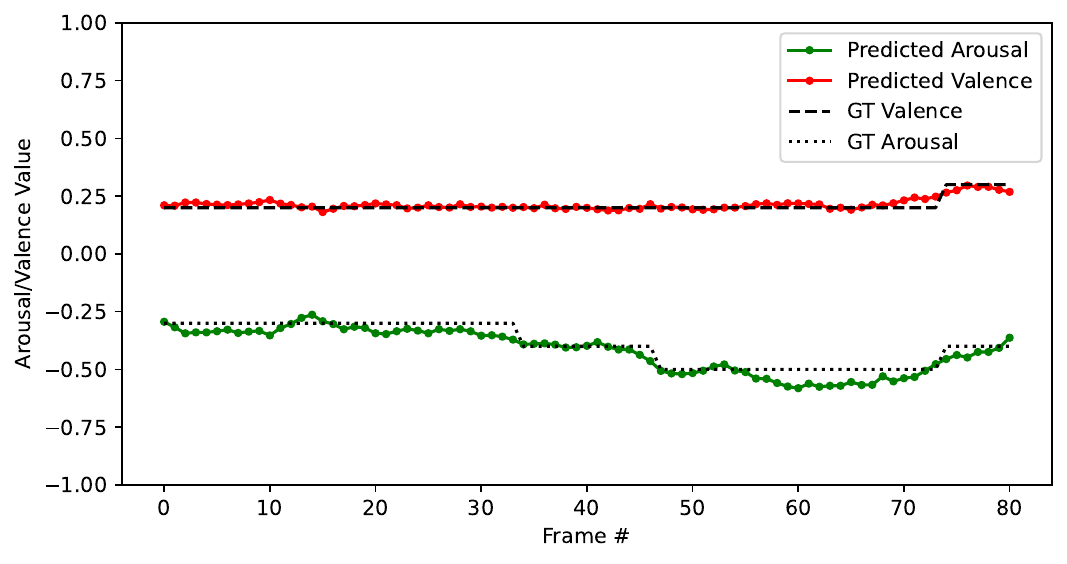}
\caption{Qualitative samples for valence and arousal estimation on samples of the AFEW-VA dataset, obtained with the frame-based ResNet+Fusion model.}\label{fig_sample_314}
\end{figure*}

\begin{figure}[h]%
\centering
\includegraphics[width=\columnwidth]{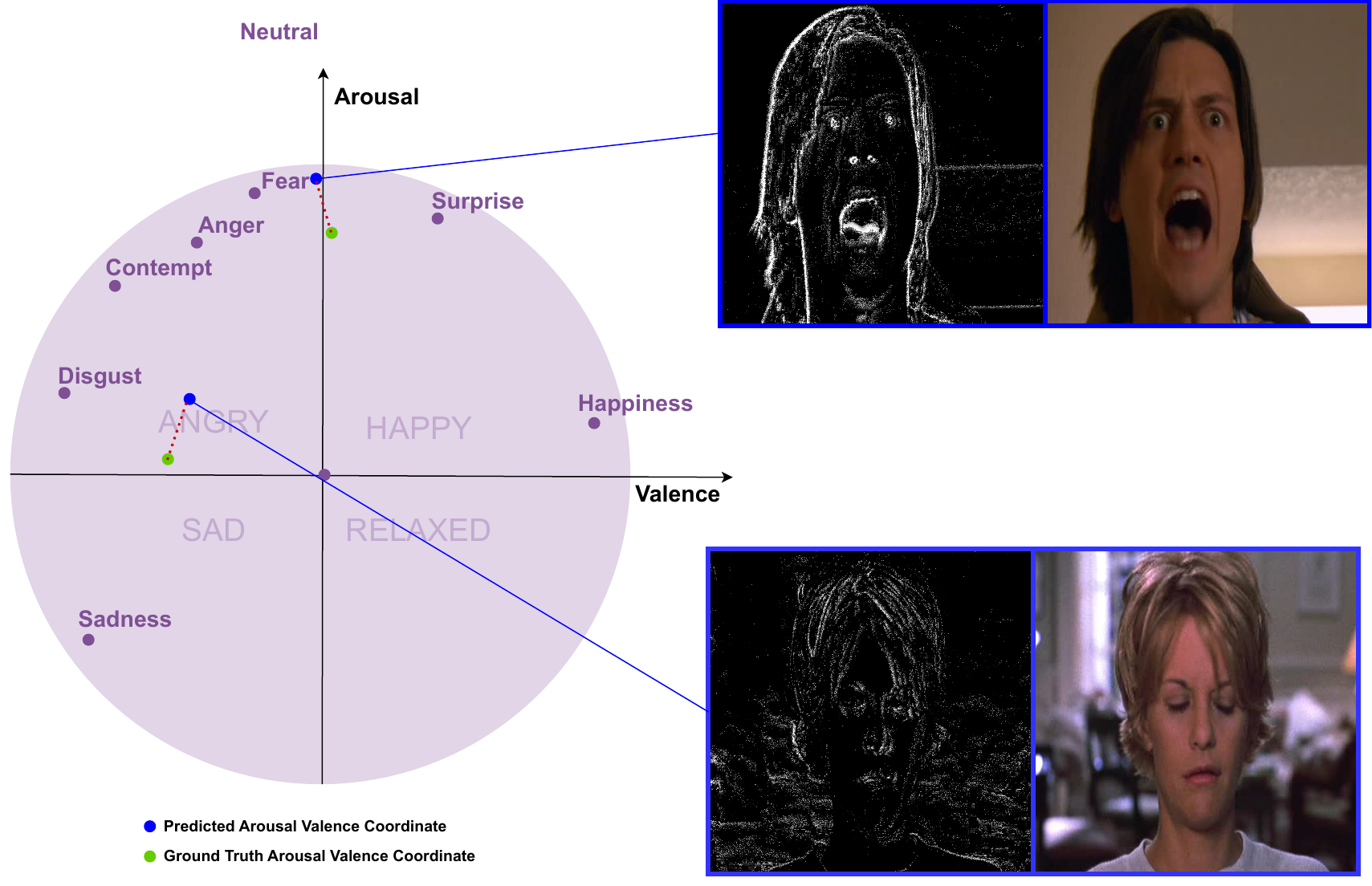}
\caption{Qualitative samples for valence and arousal estimation on samples of the AFEW-VA dataset, obtained with the frame-based ResNet+Fusion model. Estimated and ground truth valence and arousal are shown as points on the wheel of emotions.}\label{fig:sample_wheel}
\end{figure}

To provide a better understanding, we also report some qualitative results obtained with ResNet+Fusion in Fig. \ref{fig_sample_314} and Fig. \ref{fig:sample_wheel}. It can be seen that the predictions tend to adhere to the overall trend of the ground truth.

\begin{table}[t]
    \centering
    \resizebox{\linewidth}{!}{
    \begin{tabular}{lccc}
    \toprule
    Model & Modality & Arousal RMSE$\downarrow$ & Valence RMSE$\downarrow$ \\
    \midrule
    \cite{Kossaifi2017AFEWVADF} & RGB & 0.23 & 0.27 \\
    \cite{mitenkova2019valence} & RGB & 0.41 & 0.40 \\
    \cite{kollias2020deep} & RGB & 0.27 & 0.48 \\    
    \cite{handrich2020simultaneous} & RGB & 0.26 & 0.28 \\
    \cite{kossaifi2020factorized} & RGB & 0.24 & 0.24 \\
    \cite{toisoul2021estimation} & RGB & 0.22 & 0.23 \\
    \cite{parameshwara2023efficient} & RGB & 0.19 & 0.21 \\
    \midrule
    Ours (Frame) & Event & 0.17 & 0.21 \\
    Ours (Video) & Event & \textbf{0.12} & \textbf{0.19} \\
    \bottomrule
    \end{tabular}
    }
    \caption{Comparison with the state-of-the-art on AFEW-VA. }
    \label{tab:afew_comparison}
\end{table}

\begin{table}[t]
    \centering
    \resizebox{\linewidth}{!}{
    \begin{tabular}{lccc}
    \toprule
    Model & Encoding Bits & Arousal RMSE$\downarrow$ & Valence RMSE$\downarrow$ \\
        \toprule
    ResNet & 8 & \textbf{0.124} & \textbf{0.191} \\
    ViT & 8 & \textbf{0.173} & \textbf{0.211} \\
    IC3D & 8 & \textbf{0.130} & \textbf{0.201} \\
    ResNet+Transf.  & 8 & 0.133 & \textbf{0.226} \\
    \midrule
    ResNet & 16 & 0.176 & 0.218\\
    ViT & 16 & 0.232& 0.305 \\
    IC3D & 16 & 0.201& 0.302 \\
    ResNet+Transf  & 16 & \textbf{0.132} & 0.230 \\
    \bottomrule
    \end{tabular}
    }
    \caption{Comparison of different models varying the number of bits used for the event encoding strategy}
    \label{tab:tbr_bits}
\end{table}

\subsubsection{TBR ablation study}
We compare the use of different hyperparameters for the Temporal Binary Encoding. As previously stated, the natural output format for neuromorphic sensors is a continuous stream of events rather than an image. In order to leverage computer vision tools, such as CNNs, the image encoding policy plays a major role.
%As detailed in Sec. \ref{sec4}, we employ the TBR event aggregation strategy, which allows us to convert streams of events into a more canonical and recognizable form.

In this ablation study, our aim is to discern how varying the amount of bits $N$ affects the TBR encoding scheme for valence and arousal estimation. Tab. \ref{tab:tbr_bits} illustrates the outcomes for models trained on the AFEW-VA synthetic-event dataset using TBR encoding with $N=8$ and $N=16$. Notably, across all models in the table, employing an $8-bit$ encoding consistently leads to superior performance in all metrics. This phenomenon arises because using 16 bits overly compresses events, resulting in a loss of valuable information. Conversely, opting for a lower bit count, while representing a smaller number of events, leads to a more precise and informative signal, thereby facilitating superior overall performance.

\subsection{Zero-Shot Transfer on NEFER}
To establish the usefulness of training models using synthetic data, we analyze the zero-shot transfer capabilities of our models on a real event dataset.
Since there are no existing event-based datasets in the literature with annotated valence and arousal values, we use the NEFER \cite{berlincioni2023neuromorphic} dataset, which addresses the related task of emotion recognition.
%NEFER offers the task of emotion classification over the NEFER \cite{berlincioni2023neuromorphic} dataset.
Each sample is composed of an RGB video and an event stream, recorded with two separate cameras, and records the reaction from a user while being shown particular videos, chosen to trigger specific emotions. For each \textit{(user, video)} pair both the expected emotion (\texttt{A-priori}) and the one reported by the test subjects (\texttt{Reported}) are given.

In order to map frame-level valence and arousal values predicted by our models onto video-level emotions, we adopt the following approach.
We apply our model on every frame of a sequence, obtaining a temporal valence-arousal profile describing the whole video. Since samples in NEFER exhibit several static frames, with the emotion expressing itself only through extremely fast micro-movements, we choose to select only a representative frame $F$ to classify the whole video. We pick $F$ as the one with the valence-arousal pair $(F_v, F_a)$ which is farthest from the average of the sequence.
At this point, we compare $(F_v, F_a)$ against a set of prototypes corresponding to each emotion in the dataset (\textit{Disgust, Contempt, Happiness, Fear, Anger, Surprise, Sadness}). We obtain such templates $T=(T_D, T_C, T_H, T_F, T_A, T_{Su}, T_{Sa})$ by averaging the valence and arousal values estimated by our model on every frame of every video labeled with the corresponding emotion in the training set of NEFER.
The final classification $\hat{c}$ is obtained by taking the argmin of the distance between the valence-arousal pair of the reference frame and the emotion templates:
%\begin{equation}
    $\hat{c} = argmin_i  dist(F, T_i)$.
%\end{equation}
As a distance function, we use the Euclidean distance.

\begin{table}[t]
    \centering
    \resizebox{\linewidth}{!}{
    \begin{tabular}{lccc}
    \toprule
    Method & Train & Test & Accuracy \\ \midrule
    RGB~\cite{berlincioni2023neuromorphic} & NEFER & NEFER & 14.60 \\
    Event~\cite{berlincioni2023neuromorphic} & NEFER & NEFER & 22.95 \\
    Ours (Frame) & AFEW & NEFER & 19.20\\
    Ours (Video) & AFEW & NEFER & 20.80\\
    \bottomrule
    \end{tabular}
    }
    \caption{Zero-shot transfer on NEFER. We train our models on simulated events from the AFEW dataset and we test on real events from NEFER.}
    \label{tab:nefer}
\end{table}

We report the results of zero-shot transfer on NEFER in Tab. \ref{tab:nefer}.   
We show both the best performing frame-based method from Tab. \ref{tab:AFEW_frame} (ViT) and the best performing video-based method from Tab. \ref{tab:AFEW_video} (ResNet+Fusion). Interestingly, both approaches surpass the RGB baseline reported in \cite{berlincioni2023neuromorphic} in terms of classification accuracy. They also manage to achieve similar performance to the event-based model proposed in \cite{berlincioni2023neuromorphic}, i.e. a 3D convolutional network directly trained to predict emotions. This demonstrates the effectiveness of relying on simulated events for training neuromorphic models, which can then be easily deployed to work with real event data.
Note that we do not perform any additional training for the emotion classification task and we only rely on the aforementioned heuristic for inferring emotions from valence-arousal pairs.

\begin{figure}[t]%
\centering
\includegraphics[width=0.49\columnwidth]{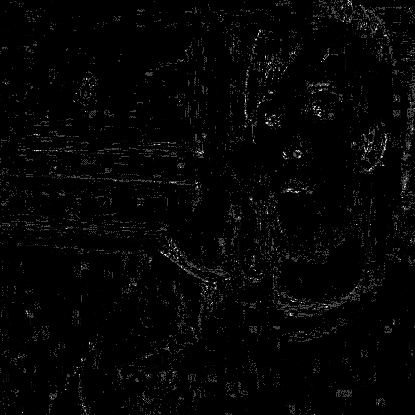}
\includegraphics[width=0.49\columnwidth]{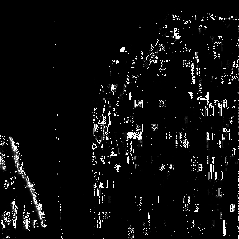}
\caption{Compression artifacts showing after postprocessing on frame samples from AFEW-VA}\label{fig:compressed}
\end{figure}

\section{Limitations and Future work}
Whilst the use of neuromorphic sensors has multiple advantages, they also have drawbacks. Mainly, these types of cameras detect local changes in brightness which means that they yield a blank frame, in case a static scene is captured, as no event is generated. This issue can be tackled with several solutions, from a simple threshold heuristic that does not update the frame unless a certain amount of events is reached, to a more sophisticated memory-equipped neural network \cite{cannici2020differentiable}.

In addition, the proposed data emulation pipeline, based on the V2E simulator, relies on good-quality input videos in order to properly approximate the event domain. In the case of heavily compressed input data, such as some of the videos in AFEW-VA, the block-sized artifacts of the MPEG compression end up as block-sized events firing synchronously (see Fig. \ref{fig:compressed}). This is in stark contrast with the real sensors that do not exhibit this type of image noise.
Such a limitation could be addressed by first restoring the original quality of the RGB frames, possibly using deep learning, e.g. GAN-based decompression frameworks \cite{galteri2017deep}.

\section{Conclusions}
In this paper, we have explored the possibility of estimating the valence and arousal of facial expressions from neuromorphic videos. To this end, we have adopted an event simulator to convert an existing RGB dataset and we have trained several models, both frame-based and video-based, on the resulting data. Interestingly, the models obtain state-of-the-art results and can also be applied zero-shot to address the downstream task of emotion recognition on real event videos, without any further training.

\small{
\subsection*{Declarations}
%All manuscripts must contain the following sections under the heading 'Declarations'.
%If any of the sections are not relevant to your manuscript, please include the heading and write 'Not applicable' for that section.
%To be used for non-life science journals

\noindent \textit{Funding}: this work was partially supported by the "Forecasting and Estimation of Actions and Trajectories for Human-robot intERactions (FEATHER)" project, funded by the University of Siena according to the PIANO PER LO SVILUPPO DELLA RICERCA (PSR 2023).
This work was partially supported by the European Commission under European Horizon 2020 Programme, grant number 951911 - AI4Media. 

\noindent \textit{Conflicts of interest/Competing interests}: none

\noindent \textit{Availability of data and material}: not applicable

\noindent \textit{Code availability}: not available
}

%\bibliographystyle{ieee_fullname}
\bibliography{sn-bibliography}% common bib file
% \section{TODO}
% \todo{

% }

\end{document}